\documentclass[twocolumn]{IEEEtran}
 
\usepackage{xspace}
\usepackage{color}
\usepackage{url}
\usepackage{graphicx}
\usepackage{amsmath}
\usepackage{amssymb}
\usepackage{hyperref}

\def\ie{i.e.\ }
\def\Deg{\ensuremath{^\circ}\xspace}
\def\Fig#1{Figure \ref{fig:#1}}
\def\Tab#1{Table \ref{tab:#1}}
\def\Sec#1{Section \ref{sec:#1}}

\begin{document}

\title{Automatic creation of urban velocity fields from aerial video}

\author{Edward Rosten, Rohan Loveland and Mark Hickman
\thanks{Edward Rosten was with Los Alamos National Laboratory. He is now
at the Engineering Department, Univeristy of Cambridge, Cambridge,  CB3 0DS, UK (email: er258@cam.ac.uk)}
\thanks{Rohan Loveland is at Los Alamos National Laboratory, 
P.O. Box 1663, Los Alamos, NM 87544, USA (email: loveland@lanl.gov)}
\thanks{Mark Hickman is at the Department of Civil Engineering and Engineering Mechanics,
University of Arizona, University of Arizona, P.O. Box 210072 Tucson, AZ  85721-0072, USA (email: mhickman@email.arizona.edu)}
}


\maketitle

\begin{abstract}

In this paper, we present a system for modelling vehicle motion in an urban scene from low frame-rate aerial video. In particular, the scene is modelled as a probability distribution over velocities at every pixel in the image.

We describe the complete system for acquiring this model.  The video is captured from a helicopter and stabilized by warping the images to match an orthorectified image of the area.  A pixel classifier is applied to the stabilized images, and the response is segmented to determine car locations and orientations.  The results are fed in to a tracking scheme which tracks cars for three frames, creating tracklets.  This allows the tracker to use a combination of velocity, direction, appearance, and acceleration cues to keep only tracks likely to be correct. Each tracklet provides a measurement of the car velocity at every point along the tracklet's length, and these are then aggregated to create a histogram of vehicle velocities at every pixel in the image.

The results demonstrate that the velocity probability distribution prior can be used to infer a variety of information about road lane directions, speed limits, vehicle speeds and common trajectories, and traffic bottlenecks, as well as providing a means of describing environmental knowledge about traffic rules that can be used in tracking.

\end{abstract}

\section{Introduction}
\label{sec:intro} 

\begin{figure}
\begin{center}
\begin{tabular}{c}
\includegraphics[width=.4\textwidth]{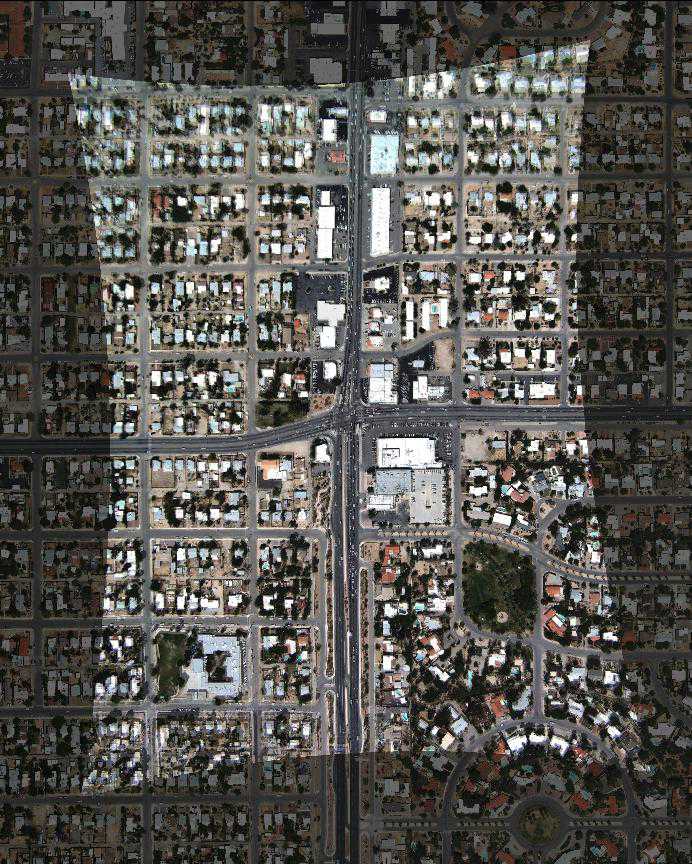}
\end{tabular}
\end{center}
\caption{\label{fig:fullField}
A video frame taken from the helicopter superimposed over an image of the
same region acquired from `Google\texttrademark\  Maps'.  The curved boundaries
of the image are due to the radial distortion correction and registration
processes.} \end{figure}

\PARstart{T}{he} increasingly widespread availability of high-resolution video sensors,
coupled with improvements in accompanying storage media, and processors
have made the collection of aerial video for the purposes of tracking
automobiles and monitoring traffic much more feasible in recent
years~\cite{Kumar01}. Corresponding efforts for the acquisition of aerial video
imagery of traffic data using helicopters for the flight platform have been
conducted by Angel et al.~\cite{Angel03}, Ernst et al.~\cite{Ernst03}, Ruh\'e et
al.~\cite{Ruhe07}, Hoogendoorn~\cite{Hoogendoorn03} and Hoogendoorn and
Schreuder~\cite{Hoogendoorn05}.  More recently, a data collect has been
conducted in a joint effort between Los Alamos National Laboratory (LANL) and
the University of Arizona, a frame of which 
is shown in \Fig{fullField}.

Aerial video is useful for traffic monitoring, since a large amount of data
can be collected without the need to manually instrument large areas with
sensors.
Therefore, it is often desirable when collecting such video data to maximize the area
covered. Limitations on storage, communications bandwidth and sensor resolution
result in trade-offs between the area and other system parameters.  In those
cases where maximum area coverage is a strong driver, the number of pixels on
target and the sampling rate will both tend to be low.  This in turn tends to
reduce tracking performance, in that virtually all tracking algorithms are
based on: 1) \textit{appearance cues}, which require enough pixels on target so
that neighboring targets are distinguishable, and/or 2) \textit{smooth
trajectory assumptions}, which are broken if sampling rates fall too low with
respect to the underlying spatial frequencies of the target trajectories.

In certain situations, tracking can be used to deduce rules underlying the
system being tracked (such as, for example occlusions~\cite{greenhill07occlusion}).
In urban environments in particular, such rules are reflected by the existence
of restricted regions (\ie cars don't drive through walls), as well as high
probabilities of certain speeds and directions of travel in allowed regions (\ie
traffic lanes have directions and speed limits). With this dataset, however we
do not need to implement a complete, fully automated  tracking system to get
these results. Instead, we use a scheme whereby we find short tracks which are
likely to be correct by ignoring difficult situations such as ambiguity, failure
to detect cars, or false positives. Since we have a large amount of data, we can
aggregate all data from all the short tracks over many frames to generate a
model representing the rules. 

In this paper, we outline an algorithm that effectively allows the inference of
these rules by building up a velocity histogram at each pixel in the scene, continuing
our previous work in~\cite{loveland_2008_improving}.
In
\Sec{data}, we describe the system for acquiring and processing data suitable
for the analysis we are performing in this paper.  This analysis is done using
track fragments, and takes advantage of the extremely high amount of information
available to allow the application of a car detector (\Sec{detection}) followed
by a tracking/data association algorithm (\Sec{tracking}) which only retains
tracks which are likely to be correct. These tracks are then used to build a
model of the city (\Sec{prior}), represented as a velocity probability
distributions at each pixel.  We then present the results in \Sec{results} and
discuss how this provides advantages not only for future tracking algorithms,
but also insight into the transportation structure of the urban environment.

\section{Data collection and initial processing}
\label{sec:data}

The data was collected in a six passenger Bell JetRanger helicopter, hovering
above Tucson, Arizona (nominally at 32.250\Deg N, 110.944\Deg W), with one door
off to allow for cables to be passed through. The sensor used was an InVision
IQEye 705 camera with the vendor's V10 optics package. The sensor was rigidly
mounted to the helicopter strut, with an ethernet interface to a laptop for
data storage and power transmission.  A marine battery was carried along in the
back seat as a power source. Details of the imaging parameters are given in
\Tab{data}.

\begin{table}
\begin{centering}
\begin{tabular}{l|r}
Parameter & Value \\\hline
Image Size & $2560\times 1920$ pixels \\
Sampling Rate & $\sim 5$ frames/second$\dagger$ \\
Spatial Resolution & $\sim 23$ cm/pixel at the centre\\
Helicopter Altitude & $\sim 900$ m (above ground level)\\
Field of view & $\sim 70\Deg$
\end{tabular}\\
\end{centering}
\caption{Data capture parameters. $\dagger$Note that the fame rate was not
entirely constant. The camera would occasionally delay frame capture by a half
or full frame interval.\label{tab:data}}
\end{table}

\begin{figure}
\includegraphics[width=0.49\textwidth]{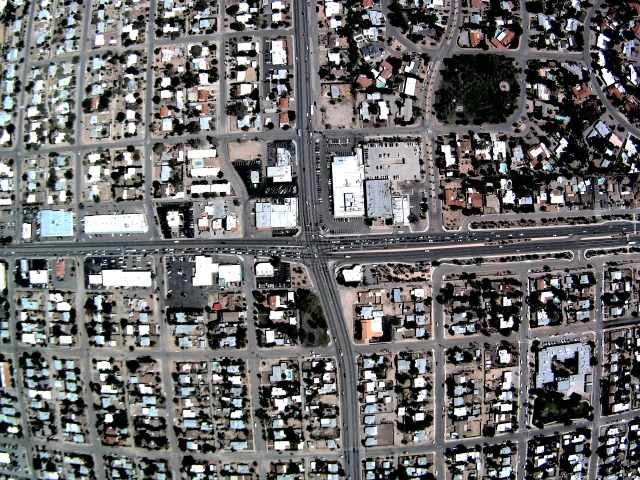}\\[1ex]
\includegraphics[width=0.49\textwidth]{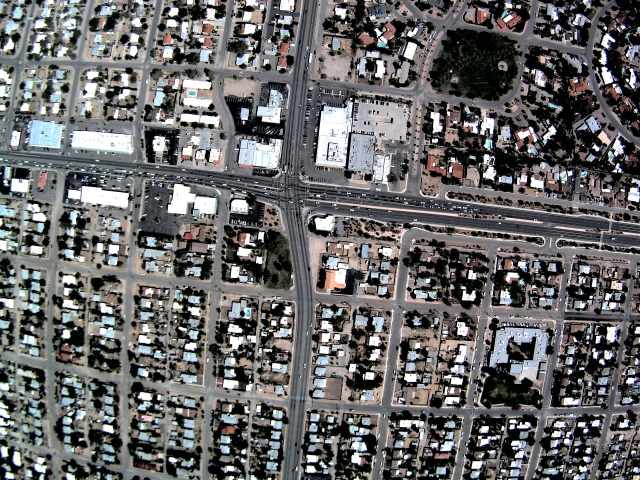}
\caption{Two frames from the aerial video dataset, taken 20s apart.. Note the
large amount of radial distortion and motion between these frames.
\label{fig:rawdata}}
\end{figure}

An example of two frames of data are shown in \Fig{rawdata}. Several features of
the data can be seen from this, namely that the camera has a large amount of
radial distortion, and that the images require stabilization. Distortions not
readily apparent from this image are due to the ground being non-planar, and the
rolling shutter on the camera. The rolling shutter is caused by pixels being
exposed shortly before the row is clocked out of the CCD, so rows towards the
bottom of the image are exposed later than rows at the top. This interacts with
the high frequency vibration of the helicopter and results in image distortion
with a high spatial frequency. This necessitates the use of a model with a
large number of parameters, so we used a robust coarse-to-fine strategy.

For a general treatment of image registration, see for
example~\cite{Brown92,Zitova03}.  In our case, registration was performed in
several stages. Firstly, the parameters of the radial distortion were found
using the method in~\cite{rosten_2008_camera}, which uses the `plumb-line
constraint' (the assumption of the existence of long, straight lines) to
determine the distortion parameters. These parameters were then used to
undistort all of the images.

Then, a single image was orthorectified by registering it to an
aerial image taken from Google Maps ({\it inter}sensor registration). An
initial, coarse registration was performed by matching SIFT~\cite{sift} features
between $\frac{1}{4}$ sized versions of the two images, and fitting a homography
using RANSAC~\cite{ransac} for robust estimation. Due to the difficulties of
matching feature points between images taken with different sensors,
this is able to fit the homography to within about a 20\% error
in scale and  about 10\Deg in rotation.

A second stage of matching was then performed using a square grid of
predetermined points, with a 50 pixel spacing between points, giving about 2000
points in total. A $75\times75$ patch of pixels was taken around each grid
point in the helicopter image. These patches were then matched to a region
around the corresponding grid point in the Google Maps image, using normalized
cross correlation. A second order polyprojective model (also known as a
rational function model~\cite{claus05rational} or nonlinear
homography~\cite{rosten06accurate}) was then fitted to these using a robust
technique.  In particular, we used a case deletion scheme where the model is
fitted to the matches by minizing the sum-squared error using the
Downhill-Simplex~\cite{downhill-simplex} algorithm.  The worst 5\% of points
are removed and the model is refitted. The procedure is iterated until the mean
error drops below 2 pixels.

The result of this is a single orthorectified helicopter image, which we refer
to as the `base image'. Subsequent images in the video are then registered to
the base image, which is {\it intra}sensor registration.  The intrasensor
registration process is inherently more forgiving than the intersensor process.
This in turn allows the usage of different subalgorithms which provide much
denser coverage of the image with better matching accuracy. The registration
processes are still similar, however, in that a coarse to fine approach is
still used, as is the basic framework of finding an optimal set of transform
parameters from an iteratively refined set of control point matches. Frames of
the video are registered simultaneously to the base image and the previous
registered image.  This removes long term drift and alleviates the problem that
the area of overlap with the base image may be small.

The first stage of the intersensor registration matches SIFT keypoints between
the current and previous/base image. An approximate transformation for the
current image is known, since we know the transformation for the previous image.
Therefore, the SIFT keypoints in the current image are only allowed to match to
points within a $50\times50$ pixel region in the previous/base image. A
polyprojective transformation is then fitted using the case deletion scheme.

After this stage, errors with high spatial frequency remain as a result of the
rolling shutter. To fit these errors, the image is split up in to $200\times200$
cells, and a displacement vector is robustly found for each cell. Intermediate
displacements are found by linearly interpolating the vectors between the cell
centres. The vectors are found by matching the SIFT keypoints between a cell in
the current image and the corresponding cell in the previous/base image. Case deletion
is then used to find the best single displacement vector for each cell.

These stages of registration bring the average jitter down to about 1 pixel, and
the registered images are used for the remainder of the paper.  For the
interested reader, a more detailed treatment of these steps is also given
in~\cite{loveland08acquisition}.

\section{Car detection}
\label{sec:detection}

A number of existing vehicle-specific techniques assume
that either a single car orientation, or a small subset of possible
orientations, is known~\cite{zhao03car,kim03fast}.
These techniques are unsuitable for our application, since cars can appear at
any orientation, for instance turning a corner at a junction. Other
algorithms~\cite{jin07vehicle,jin04vector,sharma06vehicle} use road position
information from maps to aid detection. These techniques will therefore ignore
traffic in parking lots and are not applicable where the road information is not
up to date, such as during major engineering works or evacuation situations.
Traffic monitoring would be particularly useful in these situations.  Other
techniques~\cite{moon02performance} are designed for accurate car counting,
and so are tuned to perform well in parking lots, where large number of parallel
cars are parked close together. Model based techniques, such
as~\cite{hinz03detection} tend to work with approximately 2 to 3 times the
ground sample distance of our data.  Consequently, we decided to use an approach based on
generic object detection. 

Car detection is performed by following the steps:
\begin{enumerate}
\item Apply pixel classifier to image and record the classifier response at each pixel in to a `response image'.
\item Blur the response image with a Gaussian.
\item Find all local maxima of the blurred response image.
\item Perform region growing segmentation of the response image, starting from the maxima.
\item Filter out small regions.
\item Compute mean and covariance of pixel positions in remaining regions.
\item A car is located at the mean of each region, oriented along the direction of largest covariance.
\end{enumerate}

Pixel classification is performed in a similar manner to the method presented
in~\cite{integral-image}, though somewhat simplified since processing speed is
not a concern. The centres of all cars in an image were marked. Any pixel within
6 pixels distance of a car centre was labelled as a foreground pixel. Any pixel
not within 20 pixels of any car center was labelled as a background pixel. Pixels
an intermediate distance from cars are unmarked. In our dataset, cars are
approximately 8 pixels wide and 16 pixels in length.  The background pixels are
then randomly subsampled so that foreground pixels make up about 15\% of the
data (as opposed to 1\%). A single image was marked, giving 330 cars and 300,000
labelled pixels.  We then trained an AdaBoost~\cite{adaboost} classifier on the
data for 200 iterations. 

Weak classifiers are similar to the features
in~\cite{integral-image,papageorgiou98general}, in that we use a sum of positive
and negative rectangles. In our case, we have between 1 and 5 rectangles of each
sort, and the corners of the rectangles are scatted at random with a Gaussian
distribution with $\sigma=10$ pixels. This allows the rectangles to overlap.

To detect cars, the classifier is then applied to the image. For typical
applications of machine learning, the response of the classifier would be
thresholded at zero, giving the class of each pixel. However, we wish to find
cars, rather than the class of each pixel. To do this, the classifier response
is smoothed with a Gaussian kernel (with $\sigma=3$ pixels), and all the local
maxima are found and taken as candidate car locations.

Region growing segmentation is performed iteratively from these candidate
locations until no further changes happen.  The image corresponding to the
blurred response of the classifier is denoted $C$ and the segmented image at
iteration $t$ is denoted $S_t$.  The segmentation $S_0$ is initialized such that
each pixel under a local maximum of $C$ is given a unique number, and every
other pixel has the value 0. 

The iteration sequence generates $S_t$ from $S_{t-1}$. Initially, we assign
$S_t\leftarrow S_{t-1}$.  For each occupied pixel $(x, y)$ in $S_{t-1}$, \ie
where $S_{t-1}(x, y) \neq 0$, we attempt to spread the pixel value to its
neighbours.  Consider the pixel $(x, y)$ which we attempt to spread to its
neighbour, $(x +\delta_x, y + \delta_y)$.  If $S_t(x + \delta_x, y + \delta_y) =
0$ (the neighbouring pixel is unoccupied) and if $C(x + \delta_x, y + \delta_y)
> T$ (the response is above some threshold), then assign $S_t(x + \delta_x, y +
\delta_y) \leftarrow S_{t-1}(x, y)$. We use 4-neighbour connectivity, $(\delta_x,
\delta_y)$ takes the values $(1,0), (-1,0), (0, 1)$ and $(0,-1)$.
This procedure is iterated until $S_t = S_{t+1}$.

\begin{figure*}
\begin{tabular}{cccc}
\includegraphics[width=0.5\textwidth]{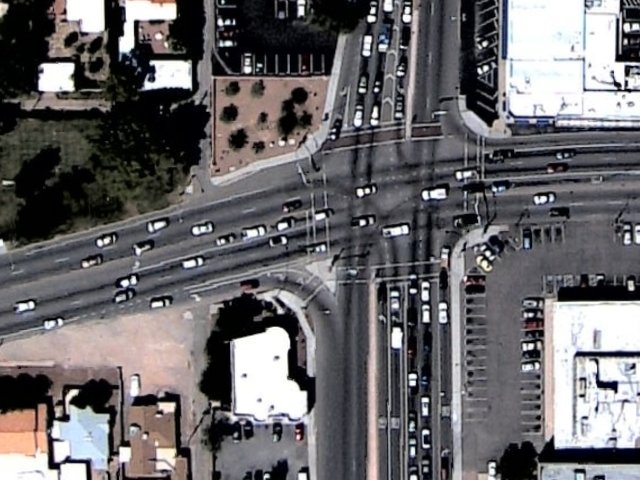}
&
\includegraphics[width=0.5\textwidth]{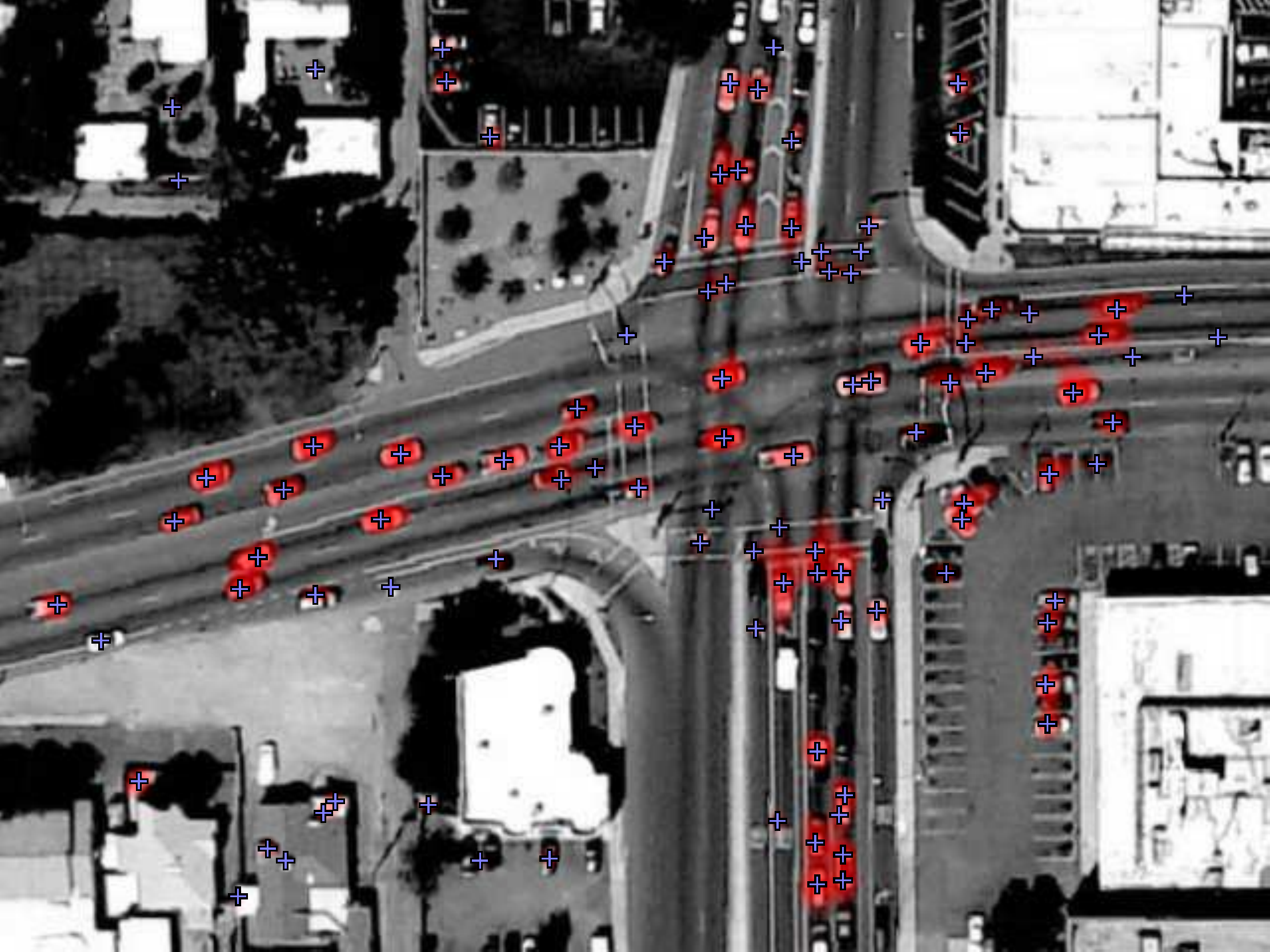}\\
1 & 2\\
\includegraphics[width=0.5\textwidth]{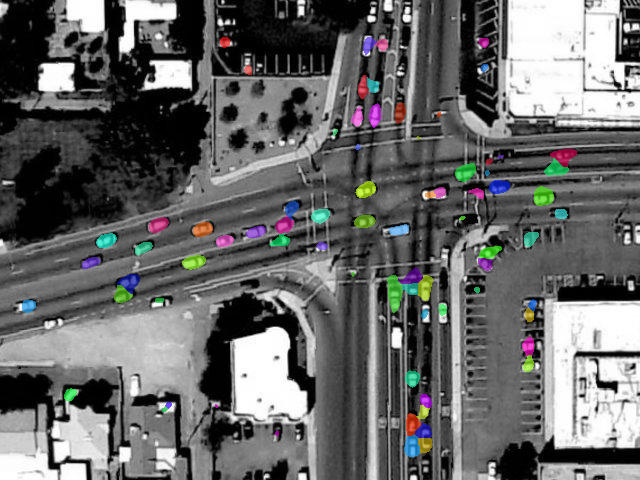}
&
\includegraphics[width=0.5\textwidth]{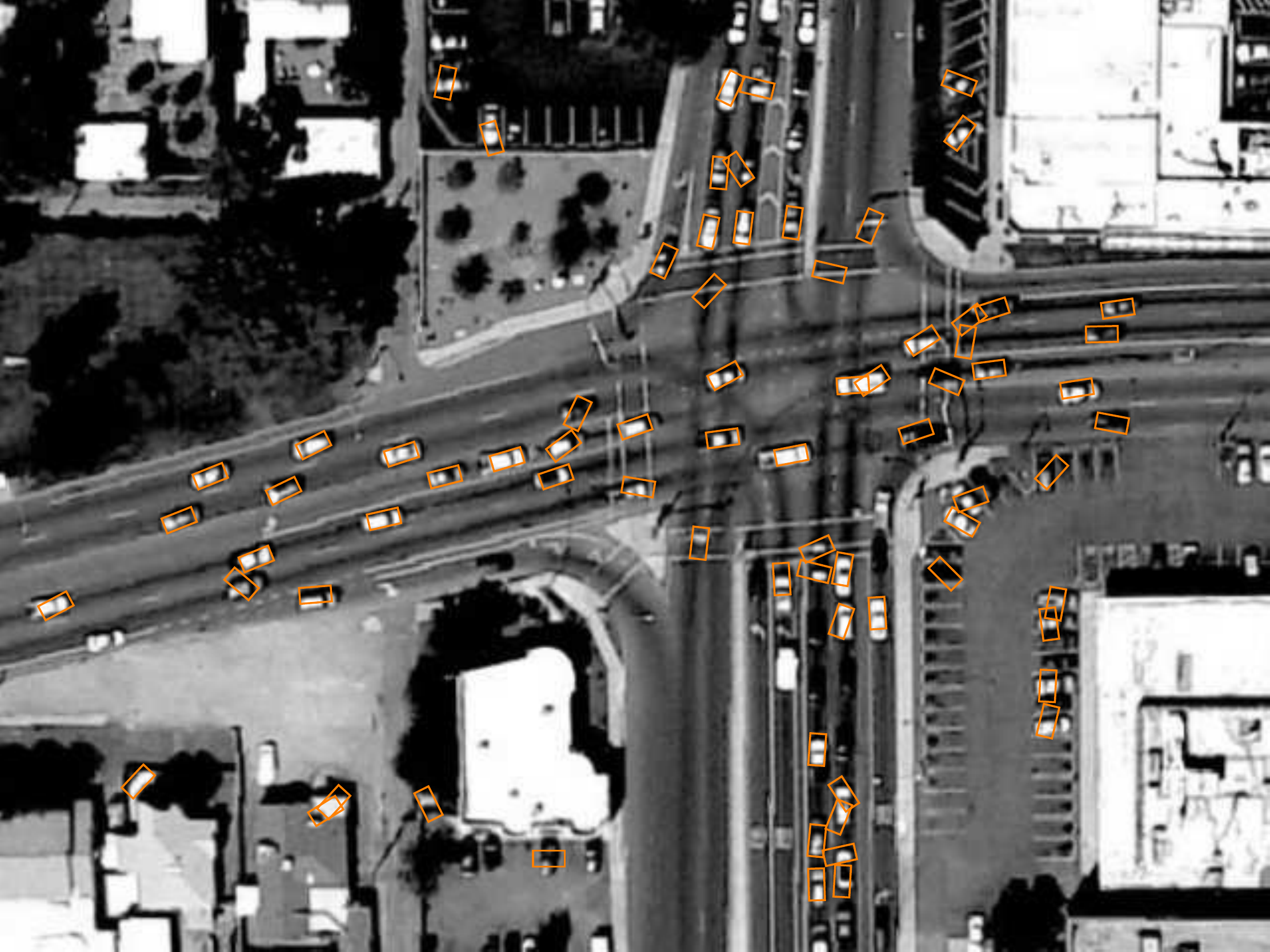}\\
3 & 4
\end{tabular}
\caption{1. Original image. Note that processing is performed in greyscale. 2.
Blurred classifier response ($C$). The classifier response (where positive) is
shown superimposed in red on the original image. Candidate car locations are
shown at the local maxima, marked by blue `+' marks. 3.  Result of grown regions
($S_\infty$). A colour was selected at random for each region.  4. The detected
cars superimposed on the original image. The rectangle representing detected
cars are drawn at a predetermined size.\label{fig:detection}}
\end{figure*}

At this point, all pixels sharing the same, nonzero number belong to the same
object. Small objects are filtered out by removing objects that do not have
enough pixels. In this case, segments with fewer than 10 pixels are removed.
Of the remaining objects, their position and orientation are found by computing
the mean and covariance of the pixels. It is assumed that cars are aligned
along the direction of largest covariance. This does not distinguish between
the front and back of cars. An illustration of this process is given in
\Fig{detection}.

\section{Tracking}
\label{sec:tracking}

There are a number of approaches to tracking multiple targets in the presence of
detection errors and measurement noise, such as JPDAF (Joint Probabilistic Data
Association Filter)~\cite{fortmann83sonar}, SMC (Sequential
Monte-Carlo)~\cite{liu98sequential} and PHD (Probability Hypothesis Density)
filter~\cite{mahler01multitarget}, which can integrate continuity assumptions
such as smoothness of motion and continuity of appearance. The main
focus of these methods is to produce long tracks which are as accurate as
possible, and get the correct data association over a large number of
measurements.

Since we wish to build up a model of car motion over the city, we can 
aggregate a large amount of data from many frames. As a result, the tracking
algorithm does not need to reliably produce long, accurate tracks. Instead we
focus on forming short, but correct track fragments. Data which makes tracking
more difficult, such as false negatives, false positives, and misestimation of
the car orientation simply causes the track to be discarded.

Without any knowledge of the system, a car in frame $n$ could match to any of
the cars in frame $n+1$.  The algorithm relies on basic physics to eliminate the
majority of potential matches. A car in frame $n$ can only match to a car
located within 30 pixels of it in frame $n+1$. Given the frame rate and pixel
size, this corresponds to a maximum velocity of 78 mph (124 km/h). Similarly,
matches where a car rotates more than 30\Deg between frames are eliminated.
Furthermore, matches where the direction of motion is more than 30\Deg off the
direction of the car are eliminated. This last constraint is not applied if the
velocity is below 5 pixels/s since noise from misregistration jitter prevents
accurate measurement of the direction of motion.

When a car in frame $n$ can still potentially match to more than one car in
frame $n+1$, we use appearance based symmetric matching~\cite{fua91combining}. That is,
for each car in frame $n$, we pick the best match in frame $n+1$, and for every
car in frame $n+1$, we pick the best match in frame $n$. Only the matches which
are consistent are retained. The quality of a match is measured by using the sum
of absolute differences (SAD) of a $8\times16$ pixel rectangular patch, centred
on a car and aligned with its direction. The best match is the one with the
smallest SAD.

The same process is then repeated between frames $n+1$ and $n+2$, and the
resulting matches are chained together to create 3 frame `tracklets'. Tracklets
with an unphysically high acceleration are removed. Although 1g corresponds to
1.74~pixels/frame/frame, we allow for an acceleration of up to
4~pixels/frame/frame to take into account jitter caused by misregistration.

\section{Model building}
\label{sec:prior}
Once the tracklets have been found over all frames, they are used to build a
model of the vehicle motion in the image. The model for motion is built up at
every pixel and contains a 2D histogram of motion vectors. The tracklets
consist of two line segments, and the two line segments each have an
associated velocity vector.  A line segment is placed in to the model by
incrementing histogram bins corresponding to the velocity at every position
along the line segments. All line segments from all tracklets are used to
create the model.  For increased precision and smoothness, instead of simply
incrementing the closest bin to the velocity vector, a smooth Gaussian blob 
centred on the velocity vector
is
entered in to the histogram.

\section{Results and Discussion}
\label{sec:results}

\begin{figure*}
\includegraphics[width=\textwidth]{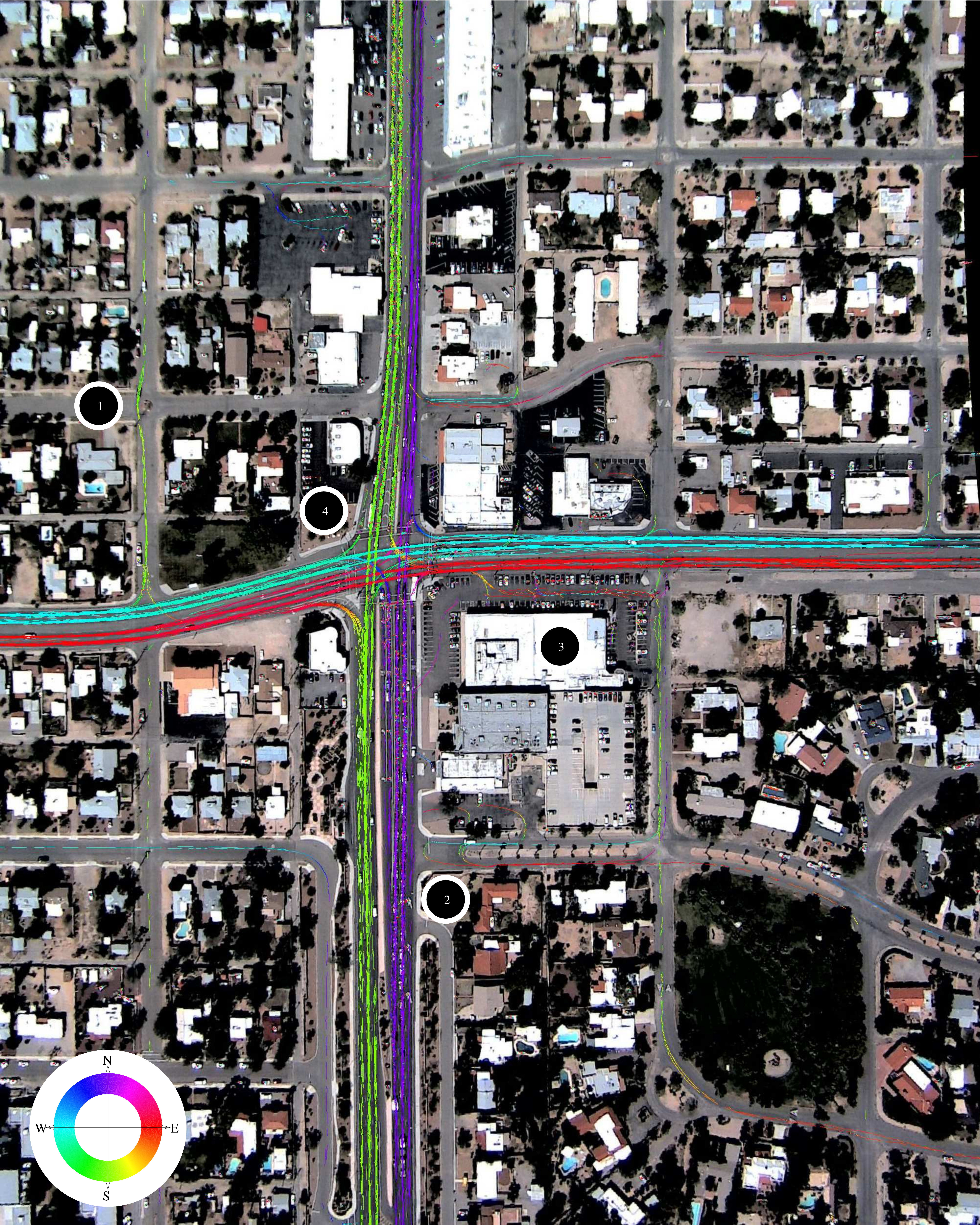}%
\caption{Direction of the mode of the histograms at each pixel,
for a cutout of the image. Where no information has been added to a histogram,
the original image is shown. The complete images have been included in the
supplementary material. Note that where the modal speed is zero, no direction
has been drawn in. The model was computed from 1126 frames (about 4 minutes
of data). Interesting parts of the data (in both images) have been labelled in
the upper image, and the complete images have been included in the supplementary
material.
\label{fig:pretty}}
\end{figure*}

\begin{figure*}
\includegraphics[width=\textwidth]{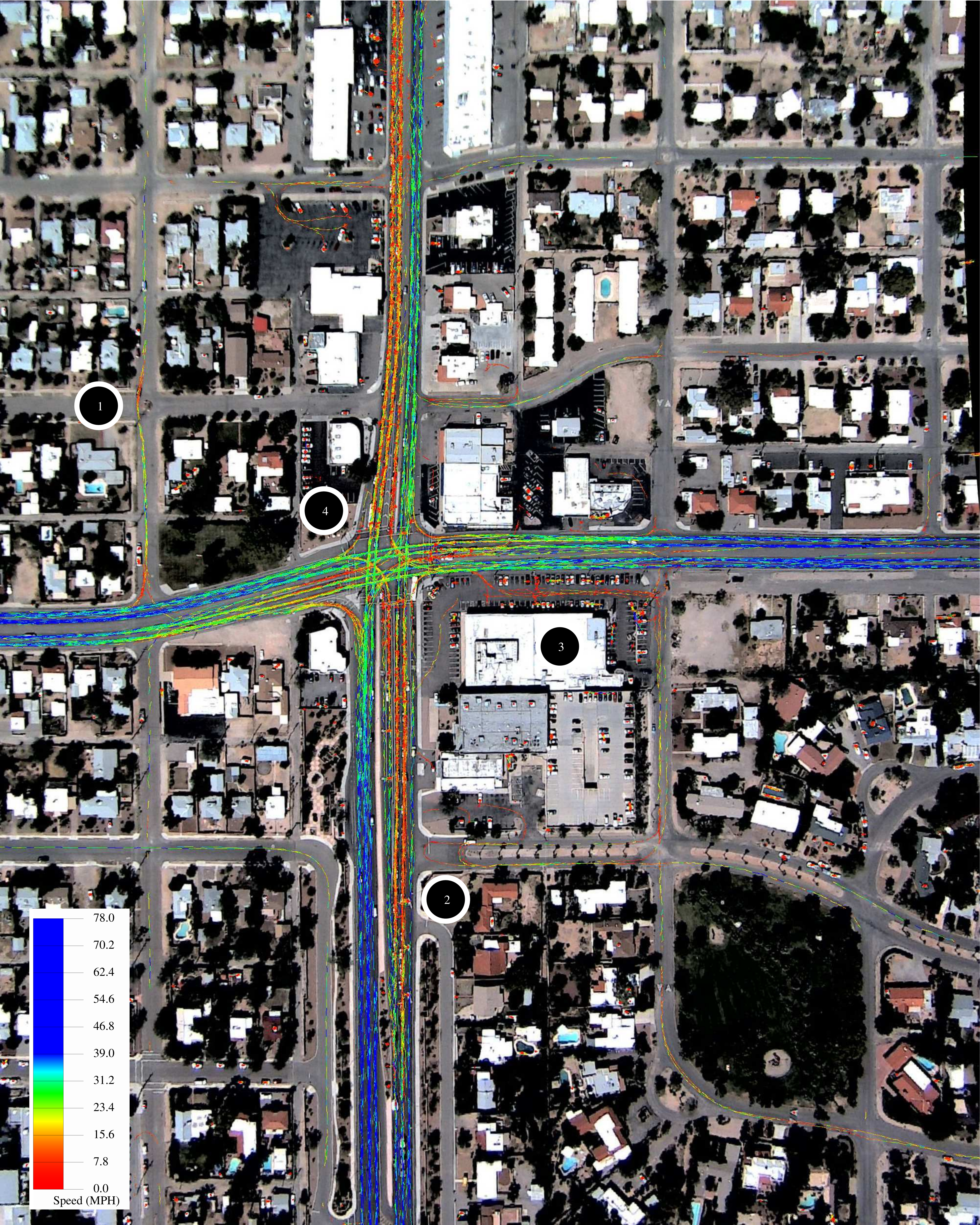}%
\caption{Speed of the mode of the histograms at each pixel,
for a cutout of the image. Where no information has been added to a histogram,
the original image is shown. The complete images have been included in the
supplementary material. Note that where the modal speed is zero, no direction
has been drawn in. The model was computed from 1126 frames (about 4 minutes
of data). Interesting parts of the data (in both images) have been labelled in
the upper image, and the complete images have been included in the supplementary
material.
\label{fig:pretty2}}
\end{figure*}

The modes of the histograms at each pixel built from a dataset are shown in
Figures \ref{fig:pretty} and \ref{fig:pretty2}. Examination of this data directly provides a large amount of
information useful for traffic engineering and planning.  Such data would be
very expensive to obtain using standard techniques. Existing loop detectors
or pneumatic tubes cannot capture vehicle trajectories; fixed cameras often
do not have an extended field of view or suffer from occlusion; and, floating
car techniques are expensive and provide only very small samples of vehicles.

Some interesting parts to note (labelled in Figures \ref{fig:pretty} and \ref{fig:pretty2}) are:
\begin{enumerate}
\item The traffic circle at this unsignalized intersection is effective and
causes a significant reduction in the speed of the vehicle. However, despite
the presence of the junction close to the south, drivers consistently accelerate
rapidly up to the nominal speed for this road.

\item A large number of U-turns at this particular junction imply that vehicles
are unwilling to enter the main northbound traffic stream at this location. There
may be significant delay for vehicles at this junction, and/or there may be
inadequate signage to deter U-turns. Note that there is a central median on the
major street which prevents traffic from turning into a southbound lane at this point.

\item  This parking lot exhibits significant traffic in the easterly direction, with
cars entering on the west and driving through to the east exit. With the proximity
to the major intersection (seen at the center of the image), it may be beneficial
to prevent cars exiting in the north-west corner, to prevent congestion and possible
conflicts at this intersection.

A failure mode of the algorithm can be seen due east of the label. The car
detector tends to misdetect cars and misestimate the orientation in parking
lots, especially when the cars are densely packed. This occasionally causes
mismatches, and therefore erroneous tracklets, the results of which can be seen.

\item The modal speed through the junction is lower than the nominal road speed.
This implies that even on a green light, there may be factors that are causing
drivers to slow, and reducing the flow of traffic. Also, the traffic speeds are
significantly lower in the north-south direction, with significant queuing and
very low speeds indicated by the tracklets. This would clearly imply that the
allocation of green time at this traffic signal may need to be adjusted to give
more green time to north-south movements.

The tracklets exiting the major intersection to the south indicate some significant
lane changing activity. This may occur in part due to potential conflicts from the
right-turning traffic from the eastbound approach to the southbound exit. This
activity may also be due in part from the expansion of the roadway from two to three
lanes as traffic moves southbound from the intersection. It may be useful to
delineate the appropriate paths of vehicles more clearly through this section of
roadway.

Similar lane changing can be noted on the eastbound exit of this intersection, but
the causes of these lane changes is not readily apparent.

\end{enumerate}

\bibliographystyle{IEEEtran}
\bibliography{papers}
\end{document}